# Improving a Multi-Source Neural Machine Translation Model with Corpus Extension for Low-Resource Languages

**Gyu-Hyeon Choi[1], Jong-Hun Shin[2], Young-Kil Kim[3]**
[1]Korea University of Science and Technology (UST), South Korea
[2]Electronics and Telecommunication Research Institute (ETRI), South Korea
[3]Electronics and Telecommunication Research Institute (ETRI), South Korea
choko93@ust.ac.kr[1], jhshin82@etri.re.kr[2], kimyk@etri.re.kr[3]

**Abstract**
In machine translation, we often try to collect resources to improve performance. However, most of the language pairs, such as Korean-Arabic and Korean-Vietnamese, do not have enough resources to train machine translation systems. In this paper, we propose the use of synthetic methods for extending a low-resource corpus and apply it to a multi-source neural machine translation model. We showed the improvement of machine translation performance through corpus extension using the synthetic method. We specifically focused on how to create source sentences that can make better target sentences, including the use of synthetic methods. We found that the corpus extension could also improve the performance of multi-source neural machine translation. We showed the corpus extension and multi-source model to be efficient methods for a low-resource language pair. Furthermore, when both methods were used together, we found better machine translation performance.

**Keywords:** Neural Machine Translation, Multi-Source Translation, Synthetic, Corpus Extension, Low-Resource

## 1. Introduction

We often try to collect resources to improve machine translation performance. Using the large size of a parallel corpus, it is possible to achieve high-quality machine translation performance. However, there are many cases where resources of language pairs are insufficient. Except for major European languages and some Asian languages, most of the language pairs do not have sufficient resources to develop a neural machine translation (NMT) system. It is also difficult to obtain parallel corpora for some language pairs such as Korean to Arabic or Korean to Vietnamese.

Since the machine translation performance largely depends on the size of a parallel corpus, it is important to find an efficient way to extend the corpus. Although it is difficult to find a proper parallel corpus, we can create an artificial parallel corpus by translating the source or target of a language pair. Some researchers have studied the extension of a parallel corpus using the pivot method (Cohn and Lapata, 2007; Utiyama and Isahara, 2007; Wu and Wang, 2007). This method introduces another language referred to as the pivot language which is a third language that is different from the source and target languages.

There are many different pivot strategies. The first is the transfer method which translates a source sentence to a pivot sentence and then to a target sentence (Cohn and Lapata, 2007; Wu and Wang, 2007). The second is the triangulation method which multiplies corresponding translation probabilities and lexical weights to create a new source-target phrase table (Utiyama and Isahara, 2007). The third is the synthetic method, which uses existing translation models to build a synthetic parallel source-target corpus from source-pivot or pivot-target (Bertoldi et al., 2008).

There are other approaches that have been proposed for multilingual training with low-resource parallel corpora. Among the approaches, there is a multi-source translation approach where the model has multiple encoders and attention mechanisms for each source language (Zoph and Knight, 2016). The goal of multi-source translation is the translation of a text given in N source languages into a single target language. This considers a case where source sentences are provided in two or more languages. In this study, we combined four other languages to achieve better target language translation. We used four source languages (Korean, English, Japanese, and Chinese) and a single target language (Arabic).

To further improve the multi-source model to be useful for low-resource language pairs, we proposed to use synthetic methods for extending a low-resource corpus and applied it to a multi-source NMT model. Although we can not obtain a high-quality corpus with these methods, it can still be effective in improving multi-source model performance.

Section 2 presents our proposed approach. Section 3 consists of the experimental settings. Section 4 contains experiment results and analysis, followed by a conclusion in section 5.

## 2. Proposed Approach

We considered a variety of ways to make a model that performs as well as an NMT model with a resource-rich corpus, even though we had to use a low-resource corpus. Among those considered, the corpus extension and multi-source translation method were employed in this study. For the corpus extension, we used a synthetic method, and there are two ways of generating the target and the source. Multi-source translation is an approach that allows one to leverage N-way corpora to improve translation quality in both resource-poor and resource-rich scenarios. Through this method, we were able to observe the improvement of machine translation performance.

### 2.1 Synthetic Method

There are two approaches to obtain a source-target parallel corpus using the source-pivot and pivot-target corpora. When we were given a pivot sentence, we translated it into a source or target sentence. In each case, translation results were combined with their source and target respectively to get a new parallel corpus. These data are referred to as the synthetic target and the synthetic source. A synthetic target is generated when a target is translated, and a synthetic source is generated when a source is translated.

### 2.1.1 Synthetic Target

The synthetic target used to obtain the target translation for source sentences in the source-pivot corpus. It can be obtained by translating pivot sentences to target sentences.

### 2.1.2 Synthetic Source

We use the synthetic source to obtain source translation for target sentences in the pivot-target corpus. It can be obtained by translating pivot sentences to source sentences. The artificial corpus created by this process is called a "synthetic source" corpus.

## 2.2 Multi-Source Translation Model

There are other approaches that have been proposed for multilingual training with low-resource parallel corpora. Among the approaches, there is the multi-source translation approach where the model has multiple encoders and attention mechanisms for each source language (Dabre et al., 2017 ; Garmash et al., 2016). Multi-source translation is the method using N source languages to improve the translation model created by using both low-resources and high-resource scenarios. This model considers a case where the source sentences are provided in two or more languages. According to this method, the model can learn more word vectors of a target language. Then the decoder will be able to generate better target sentence. In this study, we want to combine four other language pairs to get better target language translation. We used four source languages (Korean, English, Japanese, Chinese) and a single target language (Arabic). As the amount of Arabic sentences grows, the number of target word vectors will be increased. Then the word generation capability of the decoder will improve and the translation result will be better.

## 3. Experimental Settings

In this study, we used various data for the experiments, which consisted of a Korean-Arabic small-scale production parallel corpus as a baseline, and OPUS (Tiedemann et al., 2004) English-Arabic parallel corpus to make synthetic data. We used a WIT[3] (Cettolo et al., 2012) corpus to train a multi-source translation model. We used OpenNMT (Klein et al., 2017) for training the NMT systems in this study. OpenNMT is an open- source implementation of NMT that contains a library for training and deploying NMT models. To tokenize the sentences of the corpus and reduce data sparsity, we applied sub-word tokenization to the source and target sides of a training corpus with the Byte Pair Encoding (BPE) scheme (Sennrich et al., 2016). We used SentencePiece, which is an implementation of the wordpiece algorithm (Schuster and Nakajima, 2012) and BPE.

## 3.1 Languages and Data Settings

We conducted experiments with a closed production corpus (Prod), a publicly available WIT[3] corpus, and OPUS. The Prod corpus is a Korean-Arabic corpus that contains 157,865 sentences and is manually built for the traveling situation. We set the training data size of the baseline to 150,000 sentences. The WIT[3] corpus is a collection of three parallel corpora made from the transcriptions of TED (Technology, Entertainment, Design) speech, all written in the Arabic language on the target side. The language pairs of those corpora are English-Arabic, Japanese-Arabic, and Chinese-Arabic. We only used them to train the multi-source translation model (MSM). Depending on experimental, we set the training data size of each parallel corpus to 150,000 and 500,000.

| Model | Sentences |
|---|---|
| (1) Prod. Ko-Ar corpus (Baseline) | 150,000 |
| (2) (1) + Multi-Source Model (MSM) (Ko/En/Ja/Ch → Ar) | 600,000 |
| (3) (1) + Synthetic Target | 600,000 |
| (4) (1) + Synthetic Source | 600,000 |
| (5) (4) + Multi-Source Model | 2,000,000 |

Table 1: The training data size of each model.

| | Language Pair | | |
|---|---|---|---|
| WIT[3] - TED corpus | En-Ar | Ja-Ar | Ch-Ar |
| Original data size | 508,925 | 514,746 | 520,886 |
| Training data size (2) | 150,000 | 150,000 | 150,000 |
| Training data size (5) | 500,000 | 500,000 | 500,000 |

Table 2: The WIT3 data for the Multi-Source Model (MSM).

| Synthetic type | Sentences |
|---|---|
| Synthetic target (3) | 450,000 |
| Synthetic source (4) | |
| Synthetic source (5) | 350,000 |

Table 3: The synthetic corpus for using corpus extension.

To extend the training corpus, we used an OPUS English-Arabic corpus, which contains 11 million sentences, to generate a synthetic Korean-Arabic corpus. OPUS was used differently depending on whether it was used for the source side or target side. We used English as a pivot language. When a target side was created, OPUS was used to make an English-Arabic translation model. A synthetic target corpus could be obtained by translating English to Arabic. We translated English into Arabic when the given sentence existed in the Korean-English production corpus[1]. Then, we could obtain a 1.16 million parallel Korean-Arabic corpus after filtering the <unk> symbol from a 2.5 million corpus. When we manipulated the source sides, OPUS was used to obtain a good target language. It can keep Arabic language in high-quality condition. An English-Korean translation model [2] translates English sentences of an OPUS English-Arabic corpus into Korean sentences. We combined the synthetic source with the original target. Then we obtained an 800,000 Korean-Arabic parallel corpus through the filtering task. The filtering process consisted of length filtering, deduplication of sentences, and removal of sentences containing the <unk> symbol.

---
[1] This original corpus's line size is about 2.5M. The Korean-English production corpus has a trip domain.
[2] This model is an English-Korean translation model trained by ETRI.

In this paper, we used data with the sizes indicated in Tables 1, 2, and 3. From the extracted data, we selected a fixed training size. As shown in Table 2, we used a WIT[3] corpus consisting of 150,000 sentences. This is because we wanted to minimize variation of each additional corpus size in training a multi-source model. So, to train this model, we used the same size of each additional corpus with an initial baseline production corpus. Finally, we used 600,000 sentences as a multi-source corpus which consisted of Korean-Arabic (Ko-Ar), Englih-Arabic (En-Ar), Japanese-Arabic (Ja-Ar), and Chinese-Arabic (Ch-Ar) parallel language pairs. To compare fairly with the multi-source model (2) in Table 1, it is necessary to make the size of a training corpus equal. Therefore, we used 450,000 sentences of the synthetic corpus to make 600,000 sentences. When we applied the corpus extension method to a multi-source model, we set the corpus size to 500,000 sentences according to the maximum size of WIT[3]. We used the 350,000 sentence synthetic dataset to make 500,000 Korean-Arabic sentences as an initial baseline corpus. The model was trained using a total of 2 million sentences like the model (5) in Table 1.

To measure how well the model is generalizing during training, we used 3,865 development set from a Prod. We used 4,000 1-referenced test set from a Prod corpus. This test set is referred to as trip (TRIP). We extracted 2,000 Korean-Arabic sentences as a 1-referenced test set from a WIT[3] corpus. This test set is called as TED.

### 3.2 NMT and Model Settings

To train NMT systems, we used OpenNMT and we set the following conditions for training models :

- BPE vocabulary size : 8,000 vocabulary for the source language and 10,000 vocabulary for the target language in all models. When we checked the coverage of BPE models in each language, we found the appropriate size of a BPE model. This size could cover 99.5% of the words.
- Recurrent neural network (RNN) for encoders and decoders : long short-term memory (LSTM) with 4 layers, 1,000 nodes output. Each encoder is a bidirectional RNN. Word embedding size is 500 dimensions, and global attention is also enabled with default parameters.
- Optimization algorithms : stochastic gradient descent (SGD) with an initial learning rate of one which remains the same during the epoch.

We trained and evaluated the following NMT model with a WIT corpus.

- One source to one target : three models (baseline and synthetic extension corpus models)
- Four sources to one target : two models (multi-source translation models)
- Evaluate the performance of the trained models at 20 epochs.

### 3.3 Automatic Evaluations via Tokenized BLEU

We used the tokenized BLEU-4 (Papineni et al., 2002) automatic evaluation method to measure translation quality. Since Arabic is a rich-morphological language, its performance would be underestimated because non-tokenized BLEU evaluates units separated by whitespaces. Therefore, in this study, Arabic sentences were evaluated based on the results separated by morphemes. We used Farasa (Abdelali et al., 2016), which is an Arabic segmentation tool developed by the Qatar Computing Research Institute (QCRI) to tokenize Arabic words into morphemes.

## 4. Result and Analysis

### 4.1 Evaluation results

Tables 4, 5, and 6 show the BLEU scores of our proposed methods. First, we used synthetic data to determine whether the corpus extension method could improve BLEU scores. Table 4 shows the BLEU score of the model trained by a baseline corpus and the models that added synthetic data to the baseline.

For training the multi-source model, we used three different languages pairs. Table 5 showed the BLEU score when we used the multi-source model, which uses Ko-Ar, En-Ar, Ja-Ar, and Ch-Ar corpora as the training data. We found that the BLEU score is better when we use synthetic source data and the multi-source model. To gain additional improvement, we trained a multi-source model using the extended corpus by a synthetic source. Finally, based on the results, training a multi-source model with the synthetic source outperformed all other approaches in a low-resource scenario.

### 4.2 Analysis

From Tables 4,5 and 6, it is clear that we improved the quality of a translation model by using the corpus extended with a synthetic source for the multi-source model.

We have shown that the corpus extension is suitable for improving the translation model of a low-resource language pair. Table 4 shows that the BLEU score was 1.77 points higher than the baseline in the TRIP test set and 1.73 points in the TED test set when the corpus was extended to a synthetic target. However, when we used the synthetic source method, the BLEU score was increased about 4.96 and 3.86 points in the TRIP and TED test sets, respectively. Through these results, we showed that the synthetic source is more efficient in corpus extension. The reason is that generating source sentences can keep the target sentences in their original native state. The original target sentences enriched the deficient portions of a Prod corpus to improve the quality of the model. We also conducted experiments to demonstrate the effect of a multi-source model. As can be seen in Table 5, the MSM was 4.87 points higher in TRIP and 3.54 points higher in TED than the baseline. Even though the source sentences are different, the MSM can cause the model to have a lot of target information. Therefore, the model can be enhanced to obtain a better translation.

| Model | BLEU | |
|---|---|---|
| | TRIP(Prod) | TED(WIT³) |
| (1) Prod. Ko-Ar (baseline) | 21.92 | 6.19 |
| (3) (1) + Synthetic target | 23.73 | 7.92 |
| (4) (1) + Synthetic source | *26.88 | *10.05 |

Table 4: BLEU scores for the baseline and adapting extended corpus.

| Model | BLEU | |
|---|---|---|
| | TRIP | TED |
| (1) Prod. Ko-Ar (baseline) | 21.92 | 6.19 |
| (2) (1) + MSM | *26.79 | *9.73 |

Table 5: BLEU scores for the baseline and adapting multi-source model(MSM).

| Model | BLEU | |
|---|---|---|
| | TRIP | TED |
| (2) (1) + MSM | 26.79 | 9.73 |
| (4) (1) + Synthetic source | 26.88 | 10.05 |
| (5) (1) + Syn-Source + MSM | *27.07 | *12.99 |

Table 6: BLEU scores of adapting MSM and extended corpus.

Based on these results, we decided to combine the two methods. We hypothesized that the model performance would be better if we trained the extended corpus with MSM. The results are shown in Table 6. Performance was greatly improved when training a multi-source model with the synthetic source. A model obtained BLEU scores of 27.07 and 12.99 in the TRIP and TED data sets, respectively. In other words, training a multi-source model with a synthetic source can reach the improvement of 5.15 and 6.8 BLEU score for the two data sets.

## 5. Conclusion

The performance of an NMT system largely depends on the size of the parallel corpus. There are many languages in the world, but most pairs of languages are not rich enough to make a good translation model. Therefore, this paper proposed a method to improve the performance of low-resource language pairs.

In this paper, we used the corpus extension and multi-source translation method to achieve a performance improvement. The two methods of corpus extension: target generation and source generation. The source generation, called the synthetic source, can improve the performance of NMT systems. We showed the corpus extension and multi-source model to be an efficient method for low-resource languages. Furthermore, we achieved better translation performance by using both methods together.

However, the evaluation data was significantly influenced by the domain of the training data, and we found that better evaluation results were obtained in the TED evaluation than in the TRIP. If we use training data in the trip domain, we will also see a high score like the TED result. In the future, we plan to see if we can further improve the TRIP evaluation set by collecting an additional training corpus in the trip domain.

## Acknowledgements

This work was supported by Institute for Information & communications Technology Promotion(IITP) grant funded by the Korea government(MSIT)(R7119-16-1001, Core technology development of the real-time simultaneous speech translation based on knowledge enhancement)

## Bibliographical References